# The Implications of Embodiment for Behavior and Cognition: Animal and Robotic Case Studies

Matej Hoffmann and Rolf Pfeifer

Artificial Intelligence Laboratory, Department of Informatics
University of Zurich, Switzerland
{hoffmann, pfeifer}@ifi.uzh.ch

*Abstract. In this paper[1], we will argue that if we want to understand the function of the brain (or the control in the case of robots), we must understand how the brain is embedded into the physical system, and how the organism interacts with the real world. While embodiment has often been used in its trivial meaning, i.e. 'intelligence requires a body', the concept has deeper and more important implications, concerned with the relation between physical and information (neural, control) processes. A number of case studies are presented to illustrate the concept. These involve animals and robots and are concentrated around locomotion, grasping, and visual perception. A theoretical scheme that can be used to embed the diverse case studies will be presented. Finally, we will establish a link between the low-level sensory-motor processes and cognition. We will present an embodied view on categorization, and propose the concepts of 'body schema' and 'forward models' as a natural extension of the embodied approach toward first representations.*

## Introduction

Intelligent behavior has always fascinated researchers. Traditionally, intelligence was attributed solely to the control or the neural system. In 'classical' (also Good Old-Fashioned — GOFAI) Artificial Intelligence and cognitive science, the focus was on problem-solving through computation on internal symbolic representations of the world (e.g., Pylyshyn, 1987). In computational neuroscience, the focus is essentially on the simulation of certain brain regions. For example, in the 'Blue Brain' project (Markram, 2006), the focus is, for the better part, on the simulation of cortical columns — the organism into which the brain is embedded does not play a major role in these considerations. However, recently there has been an increasing interest in the notion of embodiment in all disciplines dealing with intelligent behavior, including psychology, philosophy, artificial intelligence, linguistics, and neuroscience. In this paper, we explore the far-reaching and often surprising implications of embodiment for behavior and for cognition.

---

[1] Parts of the ideas presented in this paper have appeared in previous publications; they will be referenced throughout the text.

While embodiment has often been used in its trivial meaning, i.e. 'intelligence requires a body', there are deeper and more important consequences, concerned with connecting brain, body, and environment. The behavior of any system is not merely the outcome of an internal control structure (such as the central nervous system); it is also affected by the ecological niche in which the system is physically embedded, by its morphology (the shape of its body and limbs, as well as the type and placement of sensors and effectors), and by the material properties of the elements composing the morphology. This embedding impacts the physical as well as the information (neural, control) processes that all together manifest themselves in a particular behavior (Pfeifer & Bongard, 2007).

Physical constraints shape the dynamics of the interaction of the embodied system with its environment (for example, because of the way it is attached to the body at the hip joint, during walking a leg behaves to some extent like a pendulum) and can be exploited to achieve stability and energy efficiency. We will speak about 'intelligence by mechanics' or 'morphological computation' when morphology and materials take over some of the functions normally attributed to the brain (or the control). A direct link also exists between embodiment and information: coupled sensory-motor activity and body morphology induce statistical regularities in sensory input and within the control architecture and therefore enhance internal information processing (e.g., Lungarella & Sporns, 2006).

The above-mentioned points apply to any agent interacting with its environment, animal or robot. We will present some case studies from biology, however, our selection will be biased toward case studies on robots. The advantage of using robots is that embodiment can be investigated quantitatively: robots are much simpler to manipulate and monitor. That is, first, we can change the control structure without much effort, and we can even manipulate the morphology relatively easily. Second, all sensory stimulations, motor signals, and internal states can be recorded as time series for further analysis. Having discovered some principles or put forth some hypotheses, we can turn back into the biological realm and verify the ideas. Such a method corresponds to the synthetic modeling approach, or 'understanding by building' (Pfeifer & Scheier, 1999; Webb, 2001). At the same time, these principles will enable us to design and build intelligent systems (computer programs, robots, other artifacts) for research and application purposes.

We will demonstrate that embodiment not only plays a crucial part in low-level sensory-motor activities (such as locomotion), but also in capabilities that would be considered cognitive. To illustrate that, we present an embodied view on categorization. Still, we stop short of the so-called higher-level cognitive capabilities such as planning, abstract reasoning, or language. In an effort to bridge this gap, we will sketch how the bottom-up, embodied, approach can be naturally extended to form representations, providing a way to higher-level cognition. The way is through the concepts of 'body schema' and 'forward models'.

We will proceed as follows. First, we will present a number of case studies to illustrate the physical and information theoretic implications of embodiment. The case studies have been chosen from different domains — locomotion, grasping, and visual perception — to demonstrate the broad import of the concept of embodiment. Then we will deal with the extension of the concepts toward cognition. Finally, we will attempt to integrate the diverse case studies into a general overarching scheme that captures the essence of embodiment and morphological computation, and conclude.

## Locomotion Case Studies

The fact that moving from one place to another, or locomotion, requires a body, comes as no surprise. However, it has been treated predominantly as a control problem by many; the body playing the part of a mere tool that has to be commanded appropriately. In this section, we will try to illustrate the contrary: shaping the body morphology and thereby the dynamics that result from the interaction with the environment can lead to stable and efficient locomotion, requiring very little control. We will illustrate these physical implications of being embodied on several machines and animals that walk or run. After that, a case study on leg coordination in insect walking will elucidate the impact of embodiment on information or control processes.

### Physical Implications of Embodiment in Locomotion

In this section, we want to demonstrate that the body and its dynamics in the interaction with the environment, not control, are the key determinants of locomotion behavior. First, the passive dynamic walkers — brain-less machines — will serve as a powerful illustration of this concept. Second, we will present case studies that extend this idea to powered and controlled machines. However, the goal of the brain (or controller) is not to override, but to exploit the underlying body-environment dynamics and only tune it or channel it in desired directions. We will demonstrate how such an approach leads to greater stability and energy efficiency.

**Passive dynamic walking.** The passive dynamic walker, which goes back to McGeer (1990), is capable of walking down an incline without any actuation and without control. In other words, there are no motors, no sensors, and there is no microprocessor on the robot; it is brainless, so to speak. Its locomotion is an outcome of the slope of the incline (gravity is the only power source), and the mechanical parameters of the walker (mainly leg segment lengths, mass distribution, and foot shape). The original walker had four legs to provide stability in the lateral direction; Collins et al. (2001) have constructed a two-legged version which balances by using a counter-swing of the arms that are attached rigidly to their opposing legs (see Fig. 1, A).

As the passive dynamic walkers demonstrate, locomotion can be realized through pure, but carefully tuned mechanics only. However, the 'ecological niche' (i.e. the environment in which the robot is capable of operating) is extremely narrow: it only consists of inclines of certain angles. Therefore, the next objective is to extend this concept to machines with some practical capability — that can actively walk on level ground (or even uphill) and that can cope with rough terrain.

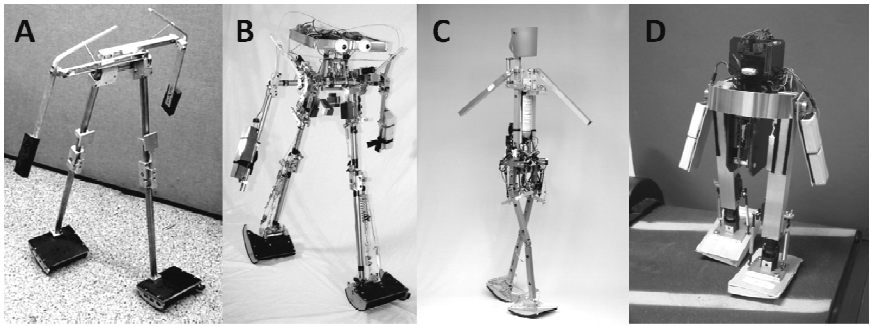

**Fig. 1. Passive dynamic and passive dynamic based walkers.** (A) The Cornell passive dynamic walker. It walks completely passively down an incline (Collins et al., 2005). (B)-(D) Passive dynamic based walkers are an extension of passive dynamic walkers. Actuation is added, such that they can walk on flat ground , but the energy-efficiency thanks to the exploitation of passive dynamics is preserved (Collins et al., 2005). (B) is an actuated extension of the passive walker (A).

**Passive dynamic based walkers.** These machines (Collins et al., 2005; Fig. 1, B-D) are a direct extension of the passive dynamic walking concept. Gravity (in the form of the incline) is substituted by small power sources. The robots can thus walk on level ground. However, they strive to preserve the advantages present in the entirely passive solution: minimal control and superior energy efficiency. The former goal can be illustrated on the Delft and Cornell bipeds that walk with simple control algorithms. Their only sensors detect ground contact, and their only motor commands are on/off signals issued once per step. The latter goal — superior energy efficiency — was also accomplished, as the cost of transport estimates testify[2].

What is the reason for the unprecedented energy efficiency of the passive dynamic based walkers? It is a consequence of the careful design of the body and of the minimalistic control scheme that only 'piggybacks' onto the underlying body dynamics. As is well known in physics, energy transfer is maximum at resonant modes of a system. The passive dynamic walkers and their active descendants contain a number of elements with pendulum-like dynamics: (1) a simple pendulum corresponds to the passive swing of the leg forward; (2) an inverted pendulum describes the motion of the hip mass over the stance leg; (3) another inverted pendulum characterizes the lateral rocking motion of the walker. The step frequency, stride length, and speed of the robots that can be observed are a direct consequence of the natural dynamics (the pendulums operating at their eigenfrequencies) that are exploited by the controller.[3]

The passive dynamic based walkers not only pave the way for energy-efficient robots of the future, but they also serve as models of human walking. The Cornell and Delft bipeds use anthropomorphic geometry and mass distributions in their legs and demonstrate ankle push-off and powered leg swinging, both present in human walking. They walk with human-like motion and human-like efficiency (Collins et al., 2005). The ease of altering different parameters and observing their effects helps us to better understand human walking.

**Self-stabilization.** Passive dynamic walkers have shown that locomotion can be realized through pure, but carefully tuned mechanics. However, how stable or adaptive is such a solution? In other words, how does a brainless machine cope with different slopes or with disturbances? The theory of nonlinear dynamical systems is often employed to analyze the phenomena involved in the mechanical (and also neural) aspects of locomotion. The walker is an example of a nonlinear dynamical system and walking patterns (which are periodic motions) correspond to limit cycles. Limit cycles in a nonlinear system can display attractive behavior, i.e. nearby trajectories are 'pulled' toward the limit cycle.

Mechanical self-stability, i.e. robustness to disturbances through local attractivity of the mechanical system, has been shown in a physical (McGeer, 1990) and mathematical (Coleman et al., 1997) walking model. In hopping or running, the dynamics is even more prolific. Fig. 2 illus-trates this phenomenon schematically. A monopod hopper driven by an open-loop controller compensates for disturbances without any explicit feedback mechanism, that is, without measuring the disturbances or altering the system. Self-stabilization has been investigated in a monopod (Ringrose, 1997), or quadruped (Poulakakis et al., 2006; Ringrose, 1997), for instance. Kubow & Full (1999) designed a dynamic model of a hexapedal runner and observed the recovery from rotational, lateral, and fore-aft velocity perturbations. Perturbations altered the translation and/or rotation of the body that consequently provided mechanical feedback by altering leg moment arms. Koditschek et al. (2004) provide an excellent review of the mechanical aspects of legged locomotion, analyzing cockroaches in particular and showing how this inspired the construction of the RHex robot — a robot with unprecedented mobility (Saranli et al., 2001). These studies show that running on rough terrain can be accomplished with simple feed-forward control in concert with a mechanical system that stabilizes passively. In the biological realm, the intrinsic properties of muscles further aid self-stability (Blickhan et al., 2007) and further assist in making the neural contribution to locomotion control simpler.

---

[2] The dimensionless mechanical specific cost of transport, $c_{mt}$ = (positive mechanical work of actuators)/(weight * distance travelled), was 0.055 for the Cornell biped, 0.08 for its Delft colleague, which is similar to the value estimated for humans (0.05), but vastly outperforms the estimated value for the state-of-the-art Honda humanoid Asimo (1.6) (Collins et al., 2005).

[3] The problem of a controller, in this case a central pattern generator, adapting to the resonant frequencies of a walking machine has been addressed by Buchli & Ijspeert, 2008 and Verdaasdonk et al., 2006.

**Body dynamics vs. control.** This confrontation is already expressed in McGeer's original paper (McGeer, 1990). The passive dynamic walker has nothing but (passive body) dynamics. On the other end of the spectrum are traditional robots with strong emphasis on control. The Honda humanoid Asimo often serves as a representative of state-of-the-art of this approach to robot locomotion. We identify the following characteristics: (1) joint trajectories are planned and enforced rather than negotiated in interaction with the environment; (2) stabilization is achieved actively (through the famous zero-moment point control scheme: Vukobratovic & Vorovac, 2004) rather than passively; (3) stiff, high-power, and high-frequency actuation is used. As a consequence of these characteristics, both computational and energetic requirements are high. On the other hand, the robot is very versatile — it can move its limbs into every possible position, it can walk uphill, downhill, even up and down the stairs.

By contrast, all the passive dynamic walker can do is walk, and it can only walk down an incline. Nevertheless, the descendants of the passive dynamics exploitation approach, the passive dynamic based walkers (Collins et al., 2005) or RHex (Saranli et al., 2001), demonstrate that the narrow ecological niche can be gradually expanded, while preserving the merits of this approach.

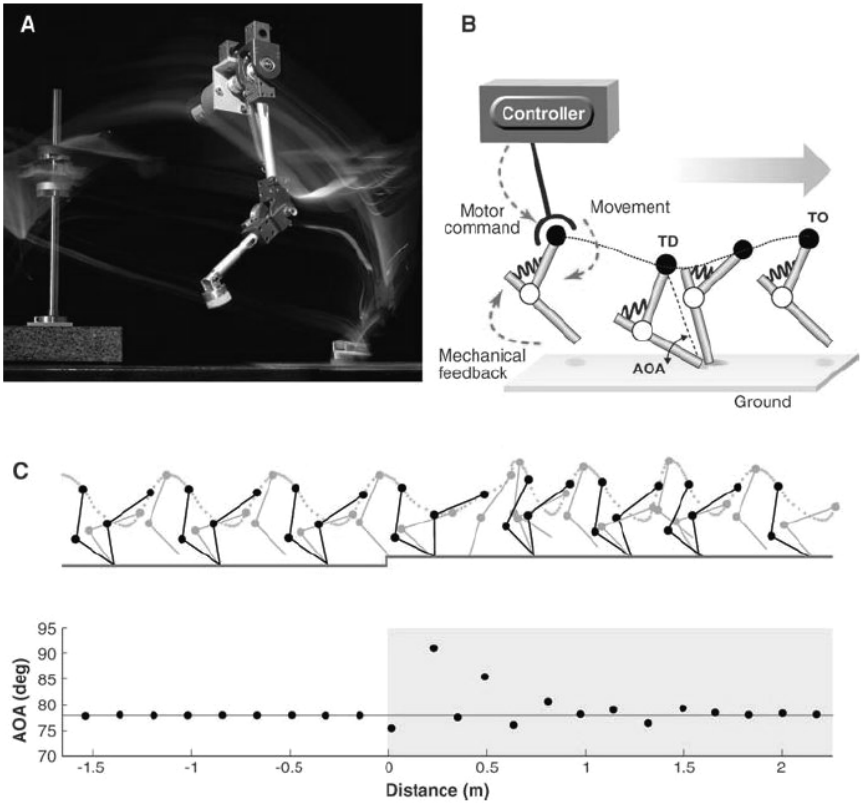

**Fig. 2. Self-stabilization.** Adaptivity is part of the mechanical structure itself. (A) Picture of a two-dimensional underactuated monoped hopping robot attached to a central rod with a rotational joint (courtesy of A. Seyfarth and A. Karguth). (B) A schematic representation of the hopping robot in the different phases of locomotion: flight, touchdown (TD) [with angle of attack (AOA)], and takeoff (TO). Only the joint depicted by the black circle (hip joint) is actuated, the knee (white circle) is passive, and the lower limb is attached to the upper limb with a simple spring. (C) Output of a simulation of the robot. The upper part of the panel shows the trajectory of the model over time as a sequence of stick figures; in the lower part, the angle of attack (the angle at which the leg hits the ground) is plotted. The model exhibits a stable hopping gait with a periodic hip motor oscillation, as indicated by the constant AOA at every step in the left side of the panel. At distance d = 0 m, there is a step in the ground that disturbs the robot's movement but to which the robot adapts without the need for any changes in the control. This purely mechanical phenomenon is called self-stabilization (Figure from Pfeifer et al., 2007; there adapted from Blickhan et al., 2007).

**Information Theoretic Implications of Embodiment in Locomotion**

The view presented in the previous section overly polarizes the situation. Body and brain should not be viewed as competitors, but rather collaborators. The tasks can be distributed and accomplished by the substrate that is more appropriate. What we have demonstrated so far is that in many locomotion-related tasks, the body itself is the candidate of choice. Nevertheless, for versatile locomotion, control is indispensable. Traditionally, control algorithms need to be fed with information about the state of the system, as obtained from sensors. Based on that, a decision, regarding the leg coordination for instance, is taken centrally. However, there are alternatives to the centralized control paradigm, which take embodiment into account. What we want to elucidate in this section is that embodiment is as important for the physical processes as it is for the informational processes. The inputs to a control scheme necessarily come through the body dynamics (see Iida & Pfeifer, 2006, for an account on sensing through body dynamics in a dynamic quadruped robot). The following case study illustrates how the body and interaction with the environment can replace a central communication between legs in insect walking.

**Leg Coordination in Insect Walking**[4]. Leg movements in insects are controlled by largely independent local neural circuits that are connected to their neighbors. There is no central controller that coordinates the legs during walking. The leg coordination comes about by the exploitation of the interaction with the environment (Cruse, 1990; Cruse et al., 2002). If the insect stands on the ground and moves forward by pushing backwards with one of its legs, as an unavoidable implication of being embodied, all the joint angles of the legs standing on the ground will instantaneously change. The insect's body is pushed forward, and consequently the other legs are also pulled forward and the joints will be bent or stretched. This fact can be exploited to the animal's advantage. All that is needed is angle sensors in the joints — and they do exist — for measuring the change, and there is global communication between the legs! But the communication is through the interaction of the agent with the environment, not through neural processing.

Inspired by the fact that the local neural leg controllers need only exploit this global communication, a neural network architecture called WalkNet has been developed which is capable of controlling a six-legged robot (Dur et al., 2003). This instance of morphological computation takes over part of the task that would have to be done by the brain — the communication between the legs and the calculation of the angles on all the joints — is performed by the interaction between the insect and the world.

---

[4] This case study has previously appeared in Pfeifer & Gomez, 2009.

## Grasping Case Studies

At first sight, grasping and locomotion do not seem to have much in common. However, as we will show in this section, the implications of embodiment illustrated thus far in locomotion can be equally well demonstrated in case studies that involve grasping. In essence, the rich and dynamic interactions of walking or running bodies with the ground will be replaced by equally complex interactions of hand morphologies and objects being grasped.

### Physical Implications of Embodiment in Grasping

In this section, we discuss how morphology and materials contribute to grasping behavior. Hand joint structure, muscle mechanics, and the distribution and density of bone to joint movements and muscle recruitment during manipulative behavior are all important variables, as investigated by Marzke & Marzke (2000). It has also been reported that ridged structure of human skin offers better grip due to increased friction (Cartmill, 1979). However, we will use two robotic case studies for our illustration of 'cheap grasping', i.e. grasping that is stable and reliable, yet requires little control. First, we will demonstrate a robotic hand, in which the attention paid to the mechanical construction leads to self-adaptation of the grasp to different objects. Second, we will present a recent universal robotic gripper, where the morphological approach was taken to its extreme.

**Cheap Grasping with a Robotic Hand[5].** The 18 degrees-of-freedom (DOF) tendon driven 'Yokoi hand' (Yokoi et al., 2004; Fig. 3) which can be used as a robotic and a prosthetic hand, is partly built from elastic, flexible, and deformable materials (this hand comes in many versions with different materials, morphologies, sensors, etc.; here we only describe one of them). The tendons are elastic, the fingertips are deformable and between the fingers there is also deformable material.

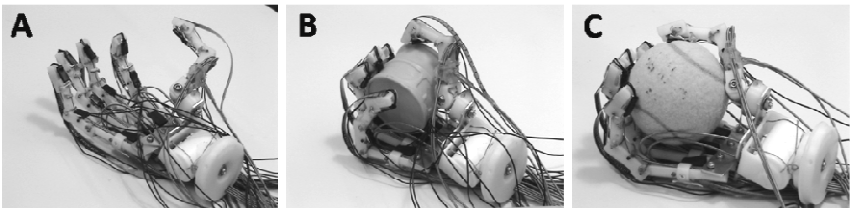

**Fig. 3: 'Cheap' grasping with a robotic hand: exploiting system-environment interaction.** (A) The Yokoi hand exploits deformable and flexible materials to achieve self-adaptation through the interaction between environment and materials. (B)-(C) Final grasp of different objects. The control is the same, but the behavior is very different.

---

[5] This case study has previously appeared in Pfeifer & Gomez, 2009.

When the hand is closed, the fingers will, because of the anthropomorphic morphology, automatically come together. For grasping an object, a simple control scheme, a 'close' is applied. Because of the morphology of the hand, the elastic tendons, and the deformable fingertips, the hand will automatically self-adapt to the object it is grasping.

**Cheap grasping with a universal gripper.** As our everyday experience confirms, a multifingered hand is an extremely dexterous manipulator. However, from a robotic perspective, this approach is highly complex from a hardware as well as software point of view. Brown et al. (2010) have therefore devised a gripper that utilizes a completely different strategy. Individual fingers are replaced by a single mass of granular material (e.g., ground coffee). The principle of operation is illustrated in Fig. 4, D. The 'bag' containing granular material is pressed onto an object, flows around it, and conforms to its shape. Then, a vacuum pump is used to evacuate air from the gripper, which makes the granular material jam and stabilize the grasp. The gripper conforms to arbitrary shapes passively, that is without any sensory feedback, thanks to its morphological properties only. Brown et al. identify three mechanism that contribute to the gripping: (i) geometric constraints from interlocking between gripper and object surfaces; (ii) static friction from normal stresses at contact; and (iii) an additional suction effect, if the gripper membrane can seal off a portion of the object's surface. The properties of the gripper can be changed by using a different granular material. Objects of various shapes (see Fig. 4, E) as well as hardness (from steel springs to raw eggs) can be gripped. An additional advantage is that the orientation of objects that are picked up and placed again does not change.

In the two case studies presented, there is no need for the agent to 'know' beforehand what the shape of the to-be-grasped object will be (which is normally the case in robotics, where the contact points are calculated before the grasping action: Molina-Vilaplana et al., 2007). In the first study, the shape adaptation is taken over by the morphology of the hand, the elasticity of the tendons, and the deformability of the fingertips, as the hand interacts with the shape of the object. In the second study, the physical properties of the granular material and how they change when air is evacuated play a key part. In both cases, control of grasping is very simple, or, in other words, very little 'brain power' is required. Clearly, these designs have their limitations; for fine manipulation more sophisticated sensing, actuation, and control may be required (Borst et al., 2002). However, a powerful fundament on which the next layers can rest has been provided.

For prosthetics, there is an interesting implication. EMG signals can be used to interface the robot hand non-invasively to a patient: even though the hand has been amputated, he or she can still intentionally produce muscle innervations which can be picked up on the surface of the skin by EMG electrodes. If EMG signals, which are known to be very noisy, are used to steer the movement of the hand, control cannot be very precise and sophisticated. But by exploiting the self-regulatory properties of the hand, there is no need for very precise control, at least for some kinds of grasping: the relatively poor EMG signals are sufficient for the basic movements (Hernandez Arieta et al., 2006; Yu et al., 2006).

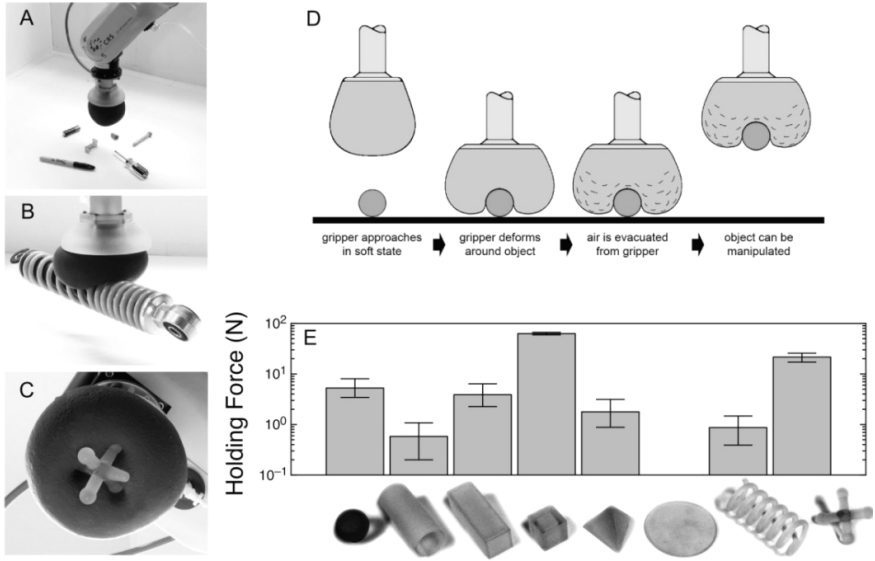

**Fig. 4. Jamming-based grippers for picking up a wide range of objects without the need for active feedback**. (A) Attached to a fixed-base robot arm. (B) Picking up a shock absorber coil. (C) View from the underside. (D) Schematic of operation. (E) Holding force Fh for several three-dimensional-printed test shapes (the diameter of the sphere shown on the very left, 2r = 25.4 mm, can be used for size comparison). The thin disk could not be picked up at all (from Brown et al., 2010, courtesy John Amend of Cornell University).

### Information Theoretic Implications of Embodiment in Grasping

As we have seen, and similarly to the locomotion case, morphology and material properties can take over a significant part of a grasping task. However, in more complex scenarios, mechanical 'intelligence' has to be aided by software or control. In order for a controller to be able to take the right decisions and issue proper motor commands, it needs to perceive the relevant information regarding the agent's interaction with the environment. Our goal in this section is to emphasize that the body morphology is as important for the perception task, as it is for taking actions. We have picked slippage sensing for our case study — a prerequisite for stable grasping and fine object manipulation — and we will show how the particular shape and material properties of an artificial skin can facilitate perception.

**Slippage detection.** In humans, the ridged skin structure not only improves the mechanics of grasping as mentioned above, but also magnifies the pressure (which can be perceived) exerted by the manipulated object (Fearing & Hollerbach, 1984), and acts as a frequency filter for specific skin mechanoreceptors (Scheibert et al., 2009). Similar properties are desirable in robotic or prosthetic hands. A wide range of tactile sensors have been developed for slippage detection which use different transduction principles: piezoelectric sensors sensitive to vibrations, skin with round ridges and strain sensors, vibrating nibs on the skin surface sensed by accelerometers, or brushes on top of capacitive membranes (see the references in Damian et al., 2010). The morphology and material properties are significantly involved in all of those designs. In what follows, we want to look in detail into yet another solution where morphology maximizes the information that can be acquired about a slippage event.

Damian et al. (2010) devised a tactile sensor consisting of a silicone skin layer with ridges a few millimeters apart which transduces surface events to a force sensing resistor beneath (Fig. 5, A). Whereas a flat skin

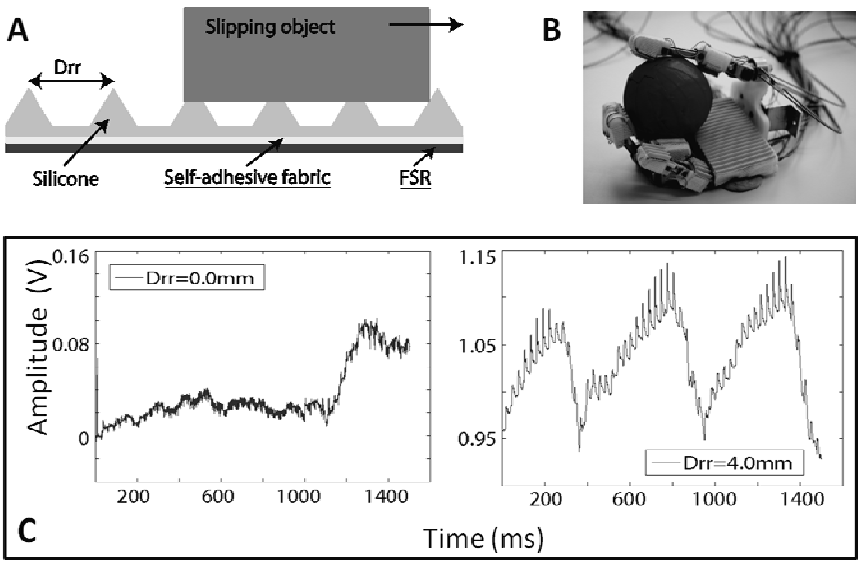

**Fig. 5: Slippage detection through ridged skin.** (A) Schematics of the artificial skin. Silicone skin with evenly spaced ridges is glued over a Force Sensing Resistor (FSR). (B) Robotic hand equipped with artificial ridged skin. (C) Signal generated by an object sliding over a skin without ridges (left), and with ridges 4 mm apart (right). The ridged skin provides a stronger signal with higher amplitude. In addition a clear periodic pattern allows for detection of slippage speed. (Damian et al., 2010)

without ridges, which was used as a reference, fails to detect an object sliding over it, ridged skin gives rise to peaks in the pressure sensor readings. Moreover, the frequency of the pressure signal obtained is directly proportional to the slippage speed and inversely proportional to the distance between ridges. The inter-ridge distance itself was found to further influence the quality of frequency encoded information. Among all skins, the one with a 4 mm spacing between ridges yielded discriminatory peak frequencies for each velocity (Fig. 5, C). The skin was afterwards employed in a robotic hand to stabilize grip. In summary, in this study, much of the electronic and algorithmic complexity present in other tactile sensing approaches has been successfully off-loaded to the morphology and allowed to detect slippage and gauge its speed with theoretically a single force sensor.

## Visual Perception Case Studies

Unlike walking or grasping, seeing seems to be concerned exclusively with perception rather than action. The goal is to acquire useful information from the environment that can be used to perform various tasks. Nevertheless, embodiment plays a key role in the information that can be acquired and such information theoretic implications of embodiment for visual perception will be the topic of this section.

A prominent theory of visual perception was proposed by David Marr (1982): vision was treated as a stage-like computational process proceeding from a two-dimensional visual array (retina/camera image) to a three-dimensional description of the world as output. Whereas this approach has lead to many successes in computer vision, robots still fall short of the capabilities that humans and animals demonstrate in object recognition, identification, and scene understanding in unstructured environments.

An alternative, and perhaps a remedy to the shortcomings of the treatment of visual perception as image processing, can be provided by embodiment. The scope of the investigation of visual perception has to be broadened to the generation of raw input image. The amount of information present in the input flow is shaped by two factors: (1) morphology of the sensory apparatus; and (2) active generation of information through sensory-motor coordination. We will address these factors separately in the sections below, but we want to stress that they always act concurrently.

Thus far, we have been referring to the information theoretic implications of embodiment in a mostly informal sense. However, the information content or structure present in the sensory and motor modalities can be quantified. Lungarella & Sporns (2006) presented several methods for measuring the (undirected) information present in sensory modalities (Shannon entropy, mutual information, integration, and complexity). To extract directed, or causal, relationships, such as from sensors to motors or vice versa, they employed transfer entropy; however, other measures are also available, as analyzed in Lungarella et al., 2007). Polani and colleagues have devised a different measure, empowerment, which measures how much influence an agent has on its environment, but only that influence that can be sensed by the agent's own sensors (see e.g., Jung et al., 2011). Yet another embodiment quantification method was presented recently by Thornton (2010), testifying the recent attention given to this subject. One of his case studies features a passive dynamic walker that we have (less formally) analyzed in the section on locomotion. Although such analysis tools are equally suited for animals and robots engaged in behavior, robots, as we have already discussed, are significantly easier to monitor and manipulate. Following the synthetic modeling approach, we will thus emphasize case studies on robots.

### The Role of Eye Morphology in Visual Perception

**Human eye.** The retina of a human eye is a variable resolution sensor: the distribution of photoreceptors is non-homogeneous. The density of cones, which are used for high acuity vision, is greatest in the center (fovea) (e.g., Curcio et al., 1990). Through this morphological arrangement, a limited number of sensing and processing elements can provide both high acuity in the center of the visual field, and a wide field of view. In robots, the retinal morphology can be emulated by the log-polar transformation (e.g., Sandini & Metta, 2002), and the degree of variable resolution can be scaled arbitrarily. Martinez et al. (2010a) investigated this effect in a robot with two eyes performing vergence behavior (simultaneous movement of both eyes in opposite directions to obtain single binocular vision). The sensor morphology as represented by the log-polar transform clearly manifests itself in the information structure calculated on a sequence of images obtained from the robot. A similar phenomenon was observed by Lungarella & Sporns (2006). There, a simulated wheeled robot (but with a human-inspired eye) was driving around colored objects and foveated on them.

**Insect eye[6].** It has been shown that for many objectives (e.g. obstacle avoidance) motion detection is all that is required. Motion detection can often be simplified if the light-sensitive cells are not spaced evenly, but if there is a non-homogeneous arrangement. For instance, Franceschini and co-workers (1992) found that in the compound eye of the house fly the spacing of the facets is denser toward the front of the animal. This non-homogeneous arrangement, in a sense, compensates for the phenomenon of motion parallax, i.e. the fact that at constant speed, objects on the side travel faster across the visual field than objects towards the front: it performs the 'morphological computation', so to speak. Allowing for some idealization, this implies that under the condition of straight flight, the same motion detection circuitry — the elementary motion detectors, or EMDs — can be employed for motion detection for the entire eye, a principle that has also been applied to the construction of navigating robots (e.g., Hoshino et al., 2000). In experiments with artificial evolution on real robots, it has been shown that certain aims, e.g. keeping a constant lateral distance to an obstacle, can be solved by proper morphological arrangement of the ommatidia, i.e. denser frontally than laterally without changing anything inside the neural controller (Lichtensteiger, 2004; Fig. 6). Because the sensory stimulation is only induced when the robot (or the insect) moves in a particular way, this is also called information self-structuring (or more precisely, self-structuring of the sensory stimulation), which leads us to the next section.

---

[6] This case study has been adapted from Pfeifer & Gomez, 2009.

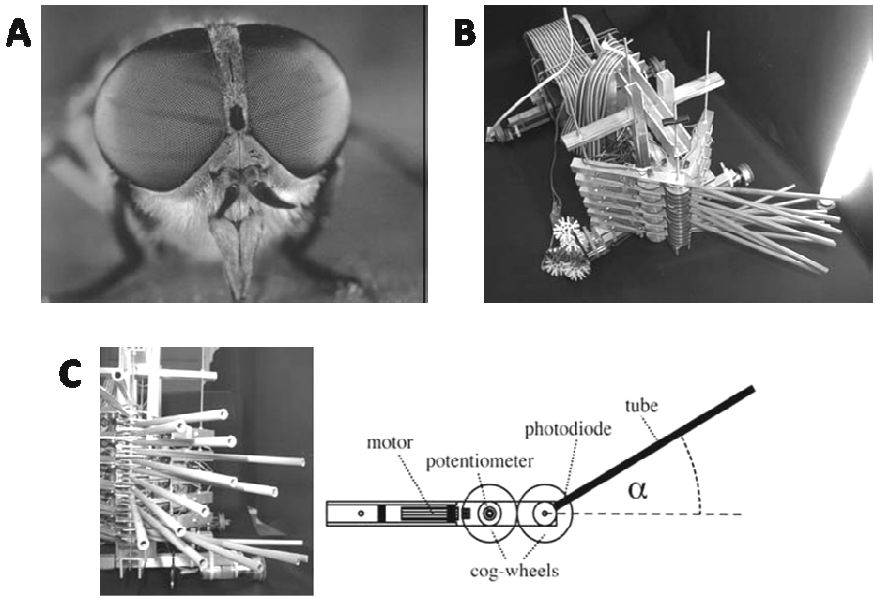

**Fig. 6. Morphological computation through sensor morphology — the Eyebot.** The specific non-homogeneous arrangement of the facets compensates for motion parallax, thereby facilitating neural processing. (A) Insect eye. (B) Picture of the Eyebot. (C) Front view: the Eyebot consists of a chassis, an on-board controller, and sixteen independently-controllable facet units, which are all mounted on a common vertical axis. A schematic drawing of the facet is shown on the right. Each facet unit consists of a motor, a potentiometer, two cog-wheels and a thin tube containing a sensor (a photo diode) at the inner end. These tubes are the primitive equivalent of the facets.

## Active Vision

The previous section has demonstrated how a particular sensor morphology affects the information structure of the raw data that reaches the sensor and that enters subsequent processing afterwards. However, the sensory stimulation is not passively received, but rather actively generated. The point we want to make was beautifully expressed by John Dewey already in 1896 (Dewey, 1896):

> We begin not with a sensory stimulus, but with a sensory-motor coordination […] In a certain sense it is the movement which is primary, and the sensation which is secondary, the movement of the body, head, and eye muscles determining the quality of what is experienced. In other words, the real beginning is with the act of seeing; it is looking, and not a sensation of light.

Only much later was Dewey's visionary observation picked up by research in active perception (e.g. Bajcsy, 1988; Churchland et al., 1994; Gibson, 1979; Noe, 2004).

Again, we will pick a robotic case study to illustrate this point. Lungarella & Sporns (2006) used an upper torso humanoid robot (Fig. 7, A) to evaluate the contribution of sensory-motor coupling to different informational measures by comparing two experimental conditions. In both conditions, the robot arm was following a preprogrammed trajectory. The movement of the ball results in a displacement of the ball relative to the head and leads to physical stimulation in the head-mounted camera. In the first condition, which we will refer to as 'fov', the sensory feedback is exploited by the controller of the robot head with camera to track the end-effector (orange ball). In other words, the sensory-motor loop (Fig. 7, B) was ensuring the orange ball stays at the center of the visual field — the fovea. In the second condition, 'rnd', the movement of the camera is unrelated to the movement of the ball (sensory-motor coupling is disrupted). The amount of information in the sequence of camera images was measured for both conditions (Fig. 7, C). As can be seen, there is more information structure in the case of the foveation condition for all measures; for example, the dark region in the center of the entropy panel indicates that entropy is clearly diminished in the center of the visual field (disorder has been reduced, or in other words, information structure has been induced), which is due to foveation being a sensory-motor coordinated behavior. Similar results were reported by Martinez et al. (2010a), who used a head with two cameras. In their case, coordinated behavior consisted in vergence, i.e. both eyes tracking salient objects. Moreover, Martinez et al. (2010a) also showed that it is not arbitrary coordinated behavior that generates information structure. A different behavior, one eye tracking the object and the other following its movements, i.e. without vergence, did not generate more information structure than random behavior. Although this behavior may seem sensory-motor coordinated to the outside observer, it does not match the robot's morphology, in this case the sensory apparatus. This illustrates the point that morphology and active perception cannot be considered in isolation.

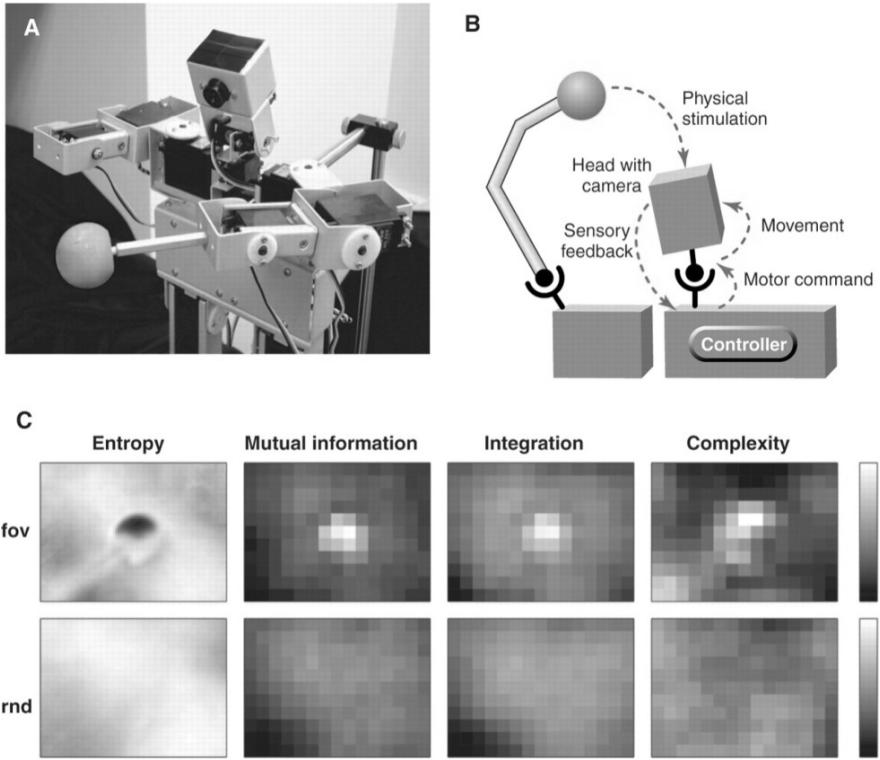

**Fig. 7. Information self-structuring.** (A) Picture of the robot, a small humanoid with a pan-tilt head equipped with a camera. (B) Schematic representation of the experimental setup. (C) Various measures to capture information structure: entropy (the amount of disorder in the system), mutual information (the extent to which the activity of one pixel can be predicted from the combined activities of neighboring pixels), integration (a measure of global coherence), and complexity (a measure that captures global coherence and local variation). The measures are applied to the camera image in the case of the foveation condition (top) and random condition (bottom). (From Pfeifer et al., 2007; there adapted from Lungarella & Sporns, 2006)

Information structure in individual sensory modalities, such as in the visual modality as shown above, is definitely a prerequisite for subsequent processing. However, for effective control of behavior we are also interested in relations between modalities, and in relations in time. In particular, we are interested in directed relations in time, such as the ones between motor and sensory modalities, which may indicate causal relations. Sensory-motor coordinated behavior increases the directed information flow, as measured using transfer entropy (Lungarella & Sporns, 2006; Martinez et al., 2010b). Such relations can be further exploited by the agent to learn to predict the consequences of its behavior. Moreover, predictability in the sensory-motor loop can be used to drive development (e.g., Oudeyer et al., 2007). Learning and representing the relations that exist between sensory and motor modalities constitute the first traces of cognition and will be the subject of the next section.

## From Sensory-motor Interaction to Embodied Cognition

Thus far, we have been dealing with relatively low-level tasks such as locomotion, grasping, or simple visual perception. We have shown that such tasks can be performed without sophisticated cognitive processing, but rather through exploitation of body dynamics and interaction with the environment. While this research is interesting in itself, how does it relate to higher-level cognition? We will provide the link in this section.

### Embodied Categorization[7]

For an autonomous embodied agent acting in the real world (e.g., an animal, a human, or a robot), perceptual categorization — the ability to make distinctions — is a hard problem (Harnad, 2005). First, based on the stimulation impinging on its sensory arrays (sensation) the agent has to rapidly determine and attend to what needs to be categorized. Second, the appearance and properties of objects or events in the environment being classified fluctuate continuously, for example owing to occlusions, or changes of distances and orientations with respect to the agent. And third, the environmental conditions (e.g., illumination, viewpoint, and background noise) vary considerably. There is much relevant work in computer vision that has been devoted to extracting scale- and translation-invariant low-level visual features and high-level multidimensional representations for the purpose of robust perceptual categorization (Riesenhuber & Poggio, 2002). Following this approach, however, categorization often turns out to be a very difficult if not an impossible computational feat, especially when sufficiently detailed information is lacking.

A solution that can only be pursued by embodied agents — but is not available when using a purely disembodied (i.e., computational) approach — is that through their interaction with the environment, agents generate the sensory stimulation required to perform the proper categorization and thus drastically simplify the problem of mapping sensory stimulation onto perceptual categories. The most typical and effective way is through a process of sensory-motor coordination. One demonstration of how sensory-motor coordination influences category formation can be found in the experiments by Pfeifer & Scheier (1997). These experiments show that mobile robots can reliably categorize big and small wooden cylinders only if their behavior is sensory-motor coordinated. A similar point is illustrated by the artificial evolution experiments of Beer (2003), where a simulated agent learns to discriminate between circular and diamond-shaped objects, or Nolfi (2002). The fittest agents, that is, those that most reliably categorized different kind of objects, were those engaging in sensory-motor coordinated behavior. Intuitively, in these examples, the interaction with the environment (a physical process) creates additional (i.e., previously absent) sensory stimulation, which is highly structured, thus facilitating subsequent information processing.

---

[7] This section has been adapted from Pfeifer et al., 2008.

Let us compare the categories that we have just come across with categories as symbols as we know them from classical symbolic AI. Taking Beer's case study, if it was realized in a symbolic architecture, we should find a 'diamond' symbol, which represents the diamonds and onto which the instances of diamonds in the real world need to be mapped (a nontrivial task, as described above). Moreover, the pitfall of this approach is that cognitive processing becomes detached from real world interaction and from meaning for the agent (the notorious symbol grounding problem: Harnad, 1990). On the other hand, when one examines the control architectures used by Pfeifer & Scheier (1997) or by Beer (2003), it is not possible to identify a site where the categories (big vs. small cylinders, or circles vs. diamonds) reside. Beer's dynamical systems analysis of the behaving agent does not reveal clear neural correlates of 'circles' or 'diamonds' either. Rather than corresponding to 'labels' defined from the outside, the categories are in fact behaviors. A small cylinder can be grasped, whereas a big one cannot; a circle is caught by the agent, whereas a diamond is avoided. Thus, categories are intrinsically meaningful to the agent and they are emergent from complex system-environment dynamics (see also Kuniyoshi et al., 2004).

On the other hand, it is probably fair to say that the discrimination tasks the agents were engaged in were of limited complexity. The opponents therefore rightly raise the question of scalability (e.g., Edelman, 2003) and argue that clearly identifiable representations allowing for hierarchical abstractions are necessary to tackle more complex scenarios. However, the dynamical systems framework and the concept of attractors that we have witnessed in the section dealing with stability in locomotion can provide a solution here. Kuniyoshi et al. (2004) or Pfeifer & Bongard (2007, ch.5), explain how, adopting the dynamical systems perspective, discretely identifiable states emerge as attractors in the combined physical and neural system of an agent. For instance, such symbols (or proto-symbols) could be gaits in a running quadruped, or they can be 'categorizing behaviors'. On top of these proto-symbols, further, more cognitive but still grounded, processing can take place.[8]

## Body Schema and Forward Models

As we have seen in the previous section, the distinction between cognitive and sensory-motor starts to blur. Categorization, perception, but even memory processes turn out to be directly coupled to sensory-motor processes and thus to embodiment (e.g., Edelman, 1987; Glenberg, 1997; Pfeifer & Scheier, 1999). What is the natural way in which an agent interacting with the world can gradually acquire cognition? We propose to follow a bottom-up and developmental pathway. Rather than starting from representations of objects or the world around the agent, we propose to start representing the very basis: the agent's body and its low-level interaction with the environment. In other words, as we have argued, any cognitive processing will always be mediated by the body and the sensory-motor loops. Therefore, these are the first candidates for an agent to learn about.

Concepts that are currently being studied, mainly in neuroscience and psychology, are 'body schema' (e.g., De Preester & Knockaert, 2005; Haggard & Wolpert, 2005; Higuchi et al., 2006; Maravita et al., 2003) and 'forward', or internal, models (Bays & Wolpert, 2007; Webb, 2004; Wolpert et al., 1998). Both concepts have also direct relevance for robotics (see e.g., Hoffmann et al., 2010, for a review). The body schema can be viewed as the sensory-motor 'representation' of the agent's body and its action possibilities. Forward models enable agents to predict the consequences of their actions and are related to anticipatory behavior (e.g., Pezzulo, 2007). In more concrete terms, for instance, in the (uncertain, dynamic, potentially hostile) world out there, it may be of advantage to: (i) predict the next sensory feedback in advance — for instance, during rapid locomotion, biological feedback is too slow; (ii) distinguish self-generated sensory information from sensory input generated by the environment, leading to detection of changes in the environment[9]; or (iii) simulate different courses of action and choose the one with the best consequences. Whereas it is not surprising that humans possess such capabilities, they have been discovered even in much simpler animals. For instance, prediction is demonstrated in the motor preparation of the prey-catching behavior of the jumping spider (Schomaker, 2004). As another example, rats are able to compare alternative paths in a T-maze before actually acting, thus 'planning in simulation' (Hesslow, 2002).

As discussed by Clark & Grush (1999), forward models are the simplest instances of circuitry that emulates the world outside and thus stands for something that is not currently present in the sensory and motor states. Thus, we may want to attribute representation to such circuitry. A 'decoupled' forward model that is not just a few steps ahead of the sensory-motor reality but that can be executed independently, in the brain only, can then be viewed as emulation/simulation of the interaction with the world, or world model. Interestingly, such a forward model can also be exploited to exercise embodied categorization, which we have presented in the previous section, in simulation. In other words, if the agent can predict the sensory consequences of its actions, it can also 'imagine' catching a circle or diamond, or grasping a cylinder. The outcome of such internal simulation can be used to derive a perceptual judgment that would otherwise not have been possible. This is demonstrated by the agent of H. Hoffmann (2007), which uses such a 'mental' rehearsal of driving in its environment to discriminate passages and dead ends.

---

[8] Maass et al., 2004 provide a neurally inspired computational model of a two-tiered architecture that could be used to implement such a processing hierarchy.

[9] For instance, it feels different when we move our eyes than when the world moves, although on the retina it may look the same.

Let us now wrap up the nature of representations and cognition that we are acquiring. Rather than representing static features (such as objects), dynamic interaction patterns, which involve the robot acting in the environment, are represented. Such representations are best viewed as motor-based. They are action-oriented, originate in the sensory-motor apparatus and remain intimately related with it (Clark & Grush, 1999; Pezzulo, 2007)[10]. Whether we want to call these phenomena 'cognitive' depends on our definition of cognition. Some views reject the cognitive/non-cognitive divide altogether, some include into the cognitive realm all kinds of adaptively valuable organism/ environment coupling (e.g., Thelen & Smith, 1994). While we consider these views equally legitimate, the view proposed by Clark & Grush (1999), among others, is that cognizers must display the capacity for environmentally decoupled thought and contemplation of options. This is exactly what a decoupled forward model provides: simulation of the world, or 'mental imagery'. This phenomenon is believed to be at the core of grounded cognition (Barsalou, 2008; Gallese & Lakoff, 2005).

## Discussion and Conclusion

We have seen a large variety of case studies. The question that immediately arises is whether there are general overarching principles governing all of them. A recently published scheme (Pfeifer et al., 2007) shows a potential way of integrating all of these ideas.

We will use Fig. 8 to summarize the most important implications of embodiment and to embed our case studies into a theoretical context.

---

[10] As opposed to symbolic AI representations that are world-centered.

Driven by motor commands, the musculoskeletal system (mechanical system) of the agent acts on the external environment (task environment or ecological niche). The action leads to rapid mechanical feedback characterized by pressure on the bones, torques in the joints, and passive deformation of skin tissue. In parallel, external stimuli (pressure, temperature, and electromagnetic fields) and internal physical stimuli (forces and torques developed in the muscles and joint-supporting ligaments, as well as accelerations) impinge on the sensory receptors (sensory system). The patterns induced thus depend on the physical characteristics and morphology of the sensory systems and on the motor commands. Especially if the interaction is sensory-motor coordinated, as in foveation, reaching, or grasping movements, information structure is generated. The effect of the motor command strongly depends on the tunable morphological and material properties of the musculoskeletal system, where by tunable we mean that properties such as shape and compliance can be changed dynamically. All parts of this diagram are crucial for the agent to function properly, but only one part concerns the controller or the central nervous system. The rest can be seen as 'morphological computation'.

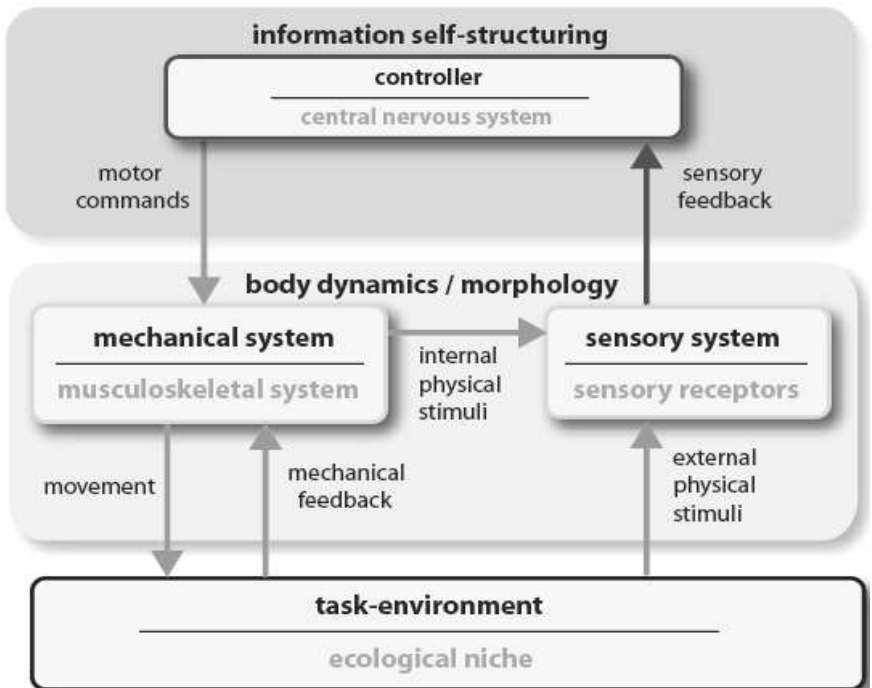

**Fig. 8: Overview of the implications of embodiment — the interplay of information and physical processes** (from Pfeifer et al., 2007; see text for details).

Let us now go through the case studies we have presented and locate them in Fig. 8. The passive dynamic walker is an instance of an interaction of the mechanical system with the environment solely — controller and sensory system are completely absent. Stabilization is achieved through the mechanical feedback loop shown in the lower left of the figure; in this case, the feedback is generated through ground reaction forces[11]. This scheme can be amended by a feed-forward controller that blindly sends motor commands to the mechanical system. That is the case for the monopod in Fig. 2 or for the hexapod RHex. As there is still no sensory system, these robots can function in the real world only thanks to mechanical self-stabilization. The 'cheap grasping' case studies illustrate a similar concept. This time, the material and morphology of the hand/gripper serve to stabilize a grasp without sensing. By contrast, the passive dynamic based walkers feature a complete scheme already — there is a sensory system and a feedback path to the controller. However, the control is rudimentary and it is still the intrinsic dynamics of the body that plays a dominant role. As a consequence of this — the intrinsic body dynamics is exploited rather than overridden — the robots also demonstrate unprecedented energy efficiency.

The study on leg coordination in insect walking provides a bridge from the physical implications of embodiment (that we have reviewed in the previous paragraph) to the information theoretic ones. Insects, when walking, also exploit mechanical feedback generated through ground reaction forces, but rather than exploiting it for gait stabilization, they capitalize on exploiting the internal sensory stimulation generated in the joint angles as one leg pushes back (thus inducing changes in the joint angles of all the other legs that are standing on the ground). This process corresponds to the lower left part of Fig. 8 and the arrow pointing from the mechanical system to the sensory system. This information can then be used for local control of individual legs. The study on slippage detection in grasping illustrates the role of the morphology of the sensory system. The particular shape of the skin — its surface is covered by ridges — magnifies the pressure exerted by objects that are grasped, and at the same time acts as a frequency filter, allowing for simply slippage speed calculation.

The case studies dealing with vision illustrate the effect of sensory morphology *and* sensory-motor coordination on the information structure that reaches a sensor. In the Eyebot, the 'insect eye' case study, given a certain behavioral pattern, e.g. moving straight, the robot induces sensory stimulation which has to be subsequently processed, for instance to achieve obstacle avoidance. The study shows that evolving a specific morphology of the facet distribution can take over a significant part of the 'processing', producing already highly structured and easy to process information for the nervous system. This process corresponds to the outer loop from the controller via mechanical system to task environment, back to sensory system and controller. The active vision case studies demonstrate the effect of action on the quality of subsequent perception, highlighting the need to treat perception as an intrinsically active process. We have also shown that the amount of sensory information can be measured quantitatively and that sensor morphology and sensory-motor coordination always go hand in hand and have to match.

---

[11] Note that the fact that the robot has no sensors and thus does not know anything about this mechanical feedback does not imply that there is no such feedback.

There are two main conclusions that can be drawn from these case studies. First, it is important to exploit the dynamics in order to achieve energy-efficient and natural kinds of movements. The term 'natural' not only applies to biological systems, but artificial systems also have their intrinsic natural dynamics. Second, there is a kind of trade-off or balance: the better the exploitation of the dynamics, the simpler the control, the less neural processing will be required. Note that all this only works, if the agent is actually behaving in the real world and therefore is generating sensory stimulation. Once again, we see the importance of the motor system for the generation of sensory signals, or more generally for perception. It should also be noted that motor actions are physical processes, not computational ones, but they are computationally relevant, or put differently, relevant for neural processing, which is why we use the term 'morphological computation'.

Having said all this, it should be mentioned that there is an additional trade-off. The more the specific environmental conditions are exploited — and the passive dynamic walker is an extreme case — the more the agent's success will be contingent upon them. Thus, if we really want to achieve brain-like intelligence, the brain (or the controller) must have the ability to quickly switch to different kinds of exploitation schemes either neurally, or mechanically through morphological change.

Finally, we have sketched a pathway how cognition can naturally emerge on top of the low-level sensory-motor processes the body is engaged in. It is the body and the interaction with the environment that are the natural candidates for first primitive representations. We want to point out that cognition is in the service of behavior here. That is, these first representations or models have to bring behavioral advantage. We have shown how this is indeed the case in simple situations where a forward model can provide an estimate of the future consequences of an action. As these simple predictive mechanisms become progressively more decoupled and autonomous, and as perhaps other processes start operating on top of them, a natural transition toward cognitive processes, which are still grounded and meaningful for the agent, has been accomplished. Therefore, unlike the original radical thesis of Brooks (1991), an embodied approach need not be anticomputationalist or anti-representationalist (Clark, 1997). Only, our view of computation and representation may have to be broadened.

## Acknowledgments

We would like to thank Dana Damian and Harold Martinez for their kind help with the preparation of the sections on grasping and visual perception respectively. We would also like to thank Igor Farkas and Keith Gunura for reviewing an earlier version of this manuscript. M. H. was supported by the Swiss National Science Foundation project "From locomotion to cognition", under Grant 200020-122279/1.